%% file: main.tex
\documentclass[10pt,twocolumn,letterpaper]{article}

\usepackage{iccv}              % To produce the CAMERA-READY version

\usepackage{comment}
\usepackage{array}
\usepackage[dvipsnames]{xcolor}
\usepackage{makecell}


\definecolor{iccvblue}{rgb}{0.21,0.49,0.74}
\usepackage[pagebackref,breaklinks,colorlinks,allcolors=iccvblue]{hyperref}

\def\paperID{2869} % *** Enter the Paper ID here
\def\confName{ICCV}
\def\confYear{2025}

\title{
Unsupervised Image Classification with Adaptive Nearest Neighbor Selection and Cluster Ensembles
}

\author{Melih Baydar$^1$ \quad Emre Akbas$^{1,2}$\\
$^1$Department of Computer Engineering, 
Middle East Technical University, Ankara, Turkey\\
$^2$Helmholtz Munich, Germany\\
{\tt\small \{melih.baydar, eakbas\}@metu.edu.tr}
}

\definecolor{ao(english)}{rgb}{0.0, 0.5, 0.0}
\definecolor{Maroon}{rgb}{0.5, 0.0, 0.0}
\DeclareMathOperator*{\argmax}{\arg\!\max}

\begin{document}
\twocolumn[{%
\renewcommand\twocolumn[1][]{#1}%
\maketitle
\begin{center}
    \centering
    \captionsetup{type=figure}
    \includegraphics[width=.95\textwidth]{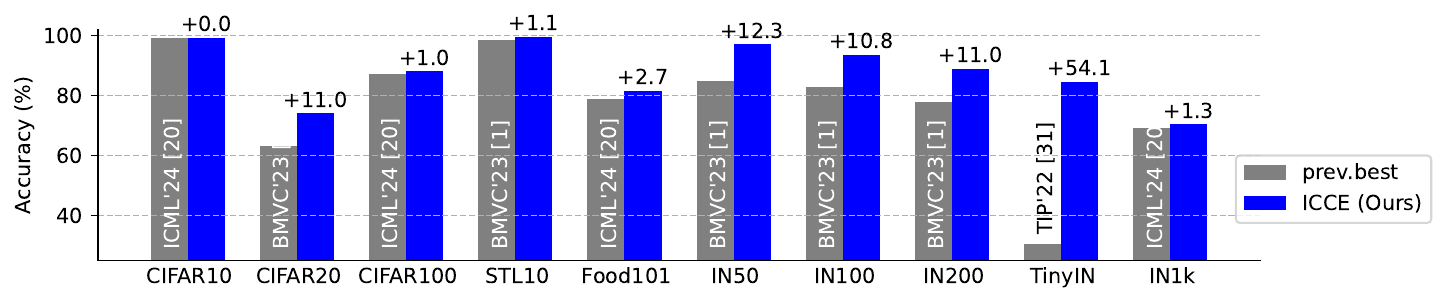}
\captionof{figure}{\textbf{Comparison with state-of-the-art.} The previous best accuracies from the literature along with our results. Our method ICCE achieves the best unsupervised image classification performance across ten datasets. Performance improvement (in terms of percentage points) of ICCE over the previous best are shown on top of blue bars.}
\end{center}%
}]
%\maketitle
\input{sec/00_abstract}    
\input{sec/01_introduction}
\input{sec/02_related_work}
\input{sec/03_method}
\input{sec/04_experiments}
\input{sec/05_conclusion}
{
    \small
    \bibliographystyle{ieeenat_fullname}
    \bibliography{references}
}

\input{X_suppl}

\end{document}

%% file: sec/00_abstract.tex
\begin{abstract}
Unsupervised image classification, or image clustering, aims to group unlabeled images into semantically meaningful categories. 
Early methods integrated representation learning and clustering within an iterative framework. 
However, the rise of foundational models have recently shifted focus solely to clustering, bypassing the representation learning step.  
In this work, we build upon a recent multi-head clustering approach by introducing adaptive nearest neighbor selection and cluster ensembling strategies to improve clustering performance. 
Our method, “Image Clustering through Cluster Ensembles” (ICCE), begins with a clustering stage, where we train multiple clustering heads on a frozen backbone, producing diverse image clusterings. 
We then employ a cluster ensembling technique to consolidate these potentially conflicting results into a unified consensus clustering.
%To unify these potentially conflicting results, we employ a cluster ensembling technique that consolidates them into a coherent consensus clustering. 
%
Finally, we train an image classifier using the consensus clustering result as pseudo-labels.
%
%Experiments on ten unsupervised image classification benchmarks show that ICCE achieves state-of-the-art performance, including 99.3\% accuracy on CIFAR10, 89.0\% on CIFAR100, and 70.4\% on ImageNet, significantly reducing the performance gap with supervised methods. 
%
%To the best of our knowledge, ICCE is the first method to exceed 70\% accuracy on ImageNet for fully unsupervised image classification, demonstrating the effectiveness of cluster ensembling at scale. 
ICCE achieves state-of-the-art performance on ten image classification benchmarks, achieving 99.3\% accuracy on CIFAR10, 89\% on CIFAR100, and 70.4\% on ImageNet datasets, narrowing the performance gap with supervised methods.
To the best of our knowledge, ICCE is the first fully unsupervised image classification method to exceed 70\% accuracy on ImageNet.
% 
%Code will be released.

\end{abstract}

%% file: sec/01_introduction.tex
\section{Introduction}

Unsupervised image classification, also known as image clustering, involves grouping unlabeled images into clusters without knowing their semantic labels.
It enables the analysis of large, unlabeled image collections %\cite{abbas2024effective}
and can serve as a precursor to tasks such as unsupervised object discovery and detection.
Early, deep learning-based approaches to unsupervised image classification performed both representation learning and clustering, iteratively repeating these steps until convergence \cite{caron2018deep, caron2020unsupervised, YM.2020Self-labelling}. 
Subsequent methods simplified this process to two stages, with the first step focused on representation learning \cite{van2020scan, adaloglou2023exploring}. 
More recently, with the rise of powerful foundational models, the representation learning step has been bypassed.
The current best unsupervised method, TEMI \cite{adaloglou2023exploring}, leverages the representations obtained by a pretrained vision transformer and trains multiple clustering heads based on these representations.
The loss function for training the clustering heads takes a training image and a randomly selected image from the set of its $k$-nearest neighbors, with the aim of maximizing the probability that these two images share the same label. 
Consequently, different clustering heads lead to different clusterings of the data, and existing work \cite{adaloglou2023exploring} only retains the head with the lowest loss, while discarding the others.

In this paper, we show that  recent methods fall short on utilizing the nearest neighbor \& multiple clustering heads  to their full extent. 
In particular, the choice of nearest neighbors to provide the self-supervision signal to the clustering heads is crucial.
We demonstrate that an adaptive $k$-nearest neighbors selection, based on distance thresholding, outperforms a fixed $k$.
Additionally, we observe that multiple clustering heads, each computing a different data grouping, capture a rich representation of relationships among images.
While TEMI \cite{adaloglou2023exploring} keeps the clustering head with the lowest cost and discards all others, we show that using all of the clustering heads through a cluster (hypergraph) ensemble method significantly improves performance. 
In addition, we apply a set of established techniques—such as output centering via Sinkhorn-Knopp—that, while well-studied in other contexts, remain largely unexplored in the domain of unsupervised image classification. These methods contribute to improved training stability and overall performance within our multi-head clustering framework.
Our experimental results show that \textbf{on ten different datasets, we set the new state of the art in fully unsupervised image classification} (see Figure 1). 

Our contributions can be summarized as follows:
\begin{itemize}
\item We propose an adaptive nearest-neighbor selection approach that better leverages the nearest-neighbor capabilities of modern self-supervised models.
\item We propose using cluster ensembling to better utilize the previously discarded multiple classifier heads.
\item We propose an enhanced unsupervised classification loss that more effectively leverages the nearest neighbors of images, enabling performance that approaches that of supervised methods across multiple benchmarks.
\item Based on these contributions, we introduce ``Image Clustering through Cluster Ensembles" (ICCE), an unsupervised image classification framework to achieve state-of-the-art performance on several benchmarks, surpassing 70\% accuracy on ImageNet, 99\% accuracy on CIFAR-10 and 88\% accuracy on CIFAR-100 datasets for the first time in fully unsupervised settings.
\end{itemize}

%% file: sec/02_related_work.tex
\section{Related Work}

\paragraph{Self-Supervised Vision Transformers.}
Vision Transformers establish a strong baseline in many computer vision tasks including image classification \citep{dosovitskiy2020image,dehghani2023scaling}, object detection \citep{carion2020end,li2022exploring,zhang2022dino} and segmentation \citep{cheng2022masked,jain2023oneformer} problems. This performance boost by ViTs also propagated to self-supervised learning \citep{caron2021emerging,zhou2021ibot,xie2022simmim,assran2022masked} due to its improved feature representation quality provided by the self-attention mechanism.

DINO \citep{caron2021emerging} explored the capabilities of ViTs with non-contrastive learning through centering and sharpening mechanism, and showed that even though training is performed with image-level objectives, ViTs also successfully learn object boundaries in the attention layers, which later motivated advancements in unsupervised localization tasks such as object discovery \citep{LOST,Wang_2022_CVPR,wang2023tokencut}, object detection \citep{wang2023cut} and image segmentation \citep{hamilton2022unsupervised,seong2023leveraging,niu2024unsupervised}. BEiT \citep{beit} and iBot \citep{zhou2021ibot} applied BERT \citep{devlin-etal-2019-bert} pretraining on ViTs, referred to as \textit{Masked Image Modeling} (MIM), where BEiT predicts the visual tokens of the original image as the learning task, while iBOT applies self-distillation on \texttt{[CLS]} tokens and patch tokens. MAE \citep{he2022masked} use an encoder-decoder architecture to predict pixels by masking a big portion of the images. In a concurrent work, SimMIM \citep{xie2022simmim} similarly propose to predict pixel values instead of patch embeddings using an $\ell_1$ loss while experimenting with various masking strategies. MSN \citep{assran2022masked} propose to apply a random mask on one augmentation of an image while leaving the other augmentation unmasked, and solve a clustering problem on the two representations using cluster prototypes. DINOv2 \citep{oquab2023dinov2} propose to curate a large-scale dataset through a self-supervised retrieval system, gathering 142M images in LVD-142M dataset. They introduce several design improvements over the iBOT method including a distillation step and obtain robust self-supervised visual features, achieving state-of-the-art k-NN and linear evaluation results on several image classification and video classification benchmark datasets.

We leverage the robust and high-quality visual representations of the DINOv2 method and propose improvements for unsupervised image classification by making better use of nearest neighbors and multi-head classifiers.

\paragraph{Unsupervised Image Classification/Clustering.}
Unsupervised image classification, also referred to as deep image clustering, has garnered significant attention as it aims to group images without relying on any human annotations. DeepCluster \citep{caron2018deep} learns feature representations through alternating between $k$-means clustering and classification phases while using the cluster assignments as pseudo-labels for training the classifier. SeLa \citep{YM.2020Self-labelling} induces an equipartition problem of the data and shows that the resulting label assignment problem is the same as the \textit{optimal transport problem}, and solves this problem with fast Sinkhorn-Knopp Algorithm rather than using the $k$-means clustering phase. SCAN \citep{van2020scan} proposed a two stage deep clustering method where they first learn a multi-head deep clustering model using nearest neighbor of images from extracted features, and then apply self-training on the pseudo-labels obtained by the model trained in the first stage. TEMI \citep{adaloglou2023exploring} extended SCAN by focusing on the clustering assignment with a novel point-wise mutual information objective while leveraging pretrained models instead of learning representations from scratch. It trains multiple clustering heads and retains only the one with the lowest cost. TURTLE \citep{gadetsky2024let} uses the representation spaces of vision foundation models and proposes an unsupervised transfer task to discover the underlying human labeling behind datasets.

We propose to further utilize the squandered multiple clustering heads in an ensembling framework to improve the clustering quality of the single classifier head during inference.

%% file: sec/03_method.tex
\begin{figure*}
\centering
\includegraphics[width=.9\linewidth]{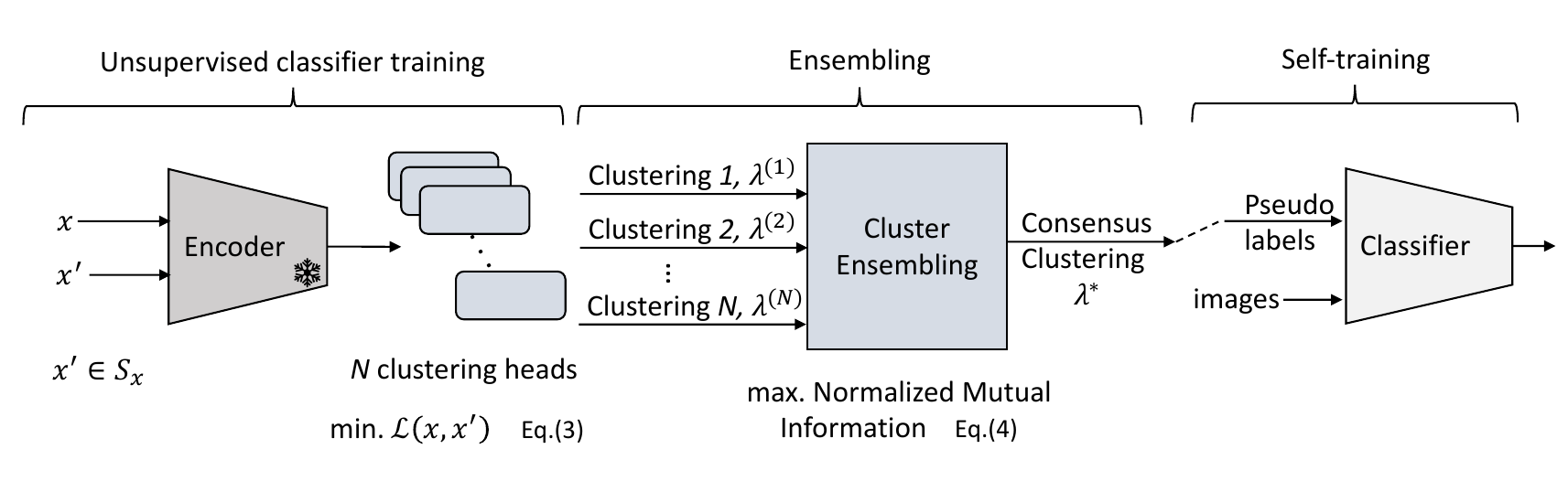}
\caption[Overview of the Training Pipeline of ICCE.]{\textbf{Overview of the training  pipeline for ICCE.}  Our method consists of three stages. 
\textbf{Stage 1 - Unsupervised classifier training:} multiple clustering heads are trained on top of the representations output by a pretrained, frozen encoder. Here $x$ is an unlabeled image and $S_x$ is the adaptively selected nearest neighbor set of $x$. Multiple heads are trained to maximize the probability of  $x$ and a randomly selected neighbor $x'$ having the same label (Equation \eqref{eq:ICCE_loss}).
\textbf{Stage 2 - Cluster ensembling:} different, potentially conflicting clusterings are unified through cluster ensembling (Equation \eqref{eq:cluster_ensemble_objective}). 
\textbf{Stage 3 - Self-training:} consensus clustering result is used as pseudo-labels and an image classifier is trained. This classifier uses the same frozen encoder from Stage 1.  \textbf{For inference,} only the classifier trained in Stage 3 is used. }
\label{fig:UCLS}
\end{figure*}

\section{Method}
In this section, we describe our proposed method, \textit{ICCE}, an unsupervised image classification approach composed of three stages. In the first stage, we provide improvements for the existing multi-head image clustering methods with two main components based on nearest neighbor utilization and training objective. Building upon the clustering outputs on the training set by the first-stage model, we employ an ensembling method to form an improved clustering of the training images in the second stage, leveraging the multiple clustering heads which previous methods typically discard. In the final stage, we perform self-training using the ensemble clustering outputs as pseudo-labels to train a classification model, which is used for inference on the test set. \textit{ICCE} achieves state-of-the-art results across several unsupervised image classification benchmarks. Figure \ref{fig:UCLS} provides an overview of \textit{ICCE}.

\subsection{Enhancing Unsupervised Multi-Head Classifier}
\label{sec:stage_1}
Current unsupervised multi-head image classification methods \cite{adaloglou2023exploring,van2020scan} provide a strong baseline for unsupervised image classification. We adopt TEMI \cite{adaloglou2023exploring} for our experiments, and explore if the self-supervised learning can be improved to train more robust multi-head classifiers by revisiting certain model and training design aspects. We systematically identify areas where these models may underperform and introduce two key components along with a bag of tricks to boost clustering model performance.

\paragraph{Adaptive Nearest Neighbors Selection.}
Through nearest neighbor analysis on foundation vision models, we observed that as the distance between two neighbors decreases, the likelihood of them sharing the same semantic label increases. While previous methods select a fixed number of nearest neighbors per sample during training, this approach may overlook low-distance, high-similarity neighbors, leading to missed opportunities for leveraging correct neighbor pairs. Building on this observation, we employed a simple yet effective distance-thresholding approach to dynamically select the nearest neighbor set per sample, maximizing the inclusion of correct pairs. To maintain a reasonable neighbor set for samples with a high average neighbor distance, we also set a minimum neighbor count, consistent with prior methods. Formally, for an image $x$, we find its nearest neighbors set $S_x$, as the group of images whose embedding vectors (extracted by the frozen backbone) are within a distance-threshold. If $|S_x|$ is less than a parameter $k$, then we keep the closest $k$ images as $S_x$. This update notably improved the clustering performance of the clustering heads without impacting the training time, as the nearest neighbor sets are precomputed before training begins. We use a distance threshold of 0.3 throughout our experiments. Our nearest neighbor analysis can be found in supplementary material.

\paragraph{Cross-Entropy Loss Term.}
To better understand the impact of the nearest-neighbor-based loss function on the clustering performance, we trained clustering heads using ground truth nearest neighbor sets, and compared the performance with linear probing, which involves training a classifier on frozen feature representations using human annotations with a cross-entropy loss function. It should be noted that human labels were not used directly during training of the clustering heads. We observed that the performance of the TEMI \cite{adaloglou2023exploring} loss on ground truth nearest neighbors does not match the level achieved by linear probing despite sharing the same frozen backbone features across both experiments. We partly attribute this discrepancy to the limitations of the TEMI loss rather than solely to the usage of nearest neighbors and the lack of human labels.

To address this issue, we take bits from the linear probing training and consider adding a cross-entropy loss term to the TEMI loss function based on the predictions $q_s(c|x)$ of the student network, and $q_t(c|x')$ of the teacher network, which is the exponential moving average (EMA) of the student network, for a sample $x$ and its neighbor $x'$. We first assign a predicted label $\hat{c}$ to the $x'$ by the argmax operation on the teacher model's output using

\begin{equation}
     \hat{c} = \argmax_{c \in C} q_t(c|x').
\end{equation}
We later use this label as the pseudo-label for  $x$ and compute the CE loss between the student network's output $q_s(c|x)$ and the one hot pseudo-label vector $\vec{\hat{c}}$, defined by

\begin{equation}
    \mathcal{L}_{CE}((q_s(c|x), \vec{\hat{c}}) = -\sum_{i=1}^{C} \vec{\hat{c}} \log(\hat{y}).
\end{equation}
With the addition of the cross-entropy loss term, the loss function becomes

\begin{equation}
    \mathcal{L}(x, x') = \mathcal{L}_{TEMI} + \lambda \mathcal{L}_{CE}((q_s(c|x), \vec{\hat{c}})
    \label{eq:ICCE_loss}
\end{equation}
where $\mathcal{L}_{TEMI}$ is the TEMI loss function, and $\lambda$ is the weighting coefficient for the cross-entropy loss term. To ensure effectiveness, we gradually increase the weight $\lambda$ of the CE loss term using a cosine schedule throughout training, because initial teacher probabilities are unreliable and improve as the training progresses. We use $\lambda = 0.5$ throughout our experiments.

\paragraph{Bag of Tricks for Multi-Head Clustering.}
We further propose to use a set of techniques to improve unsupervised multi-head classifier training process. We briefly introduce the changes that provide stability and performance improvements over the baseline.

\begin{itemize}
    \item \textbf{Enhanced Feature Space (\citet{oquab2023dinov2}).} Inspired by DINOv2’s linear evaluation protocol, we enhance our backbone features by incorporating outputs from the final attention block alongside the CLS token. These enriched features are used both for nearest neighbor mining and during the training of classification heads. This leads to higher-quality neighbors across datasets and improves clustering performance by operating in a more expressive feature space.
    \item \textbf{Batch Normalization Layer Following the Backbone.} Similar to the findings in TEMI \cite{adaloglou2023exploring}, we also note that backbone models generate unnormalized features, leading to unstable training. While TEMI paper addressed this by standardizing feature embeddings using precomputed statistics, referred to as embedding normalization, we observed a performance drop in the case of DINOv2 features. Therefore, we introduce a batch normalization layer after the backbone model to ensure a more stable training process.
    \item \textbf{Sinkhorn-Knopp centering (\citet{caron2020unsupervised}).} We adopt the approach outlined in \citet{ruan2022weighted} and follow \cite{oquab2023dinov2} to incorporate Sinkhorn-Knopp normalization onto the outputs of the teacher model within the multi-head classifier. This adjustment promotes a more uniform utilization of outputs, aligning with the classes in the multi-head classification framework. We run the Sinkhorn-Knopp algorithm for 3 iterations.
    \iffalse
    \item \textbf{Multiple Neighbors Smoothing.} To mitigate potential errors in the nearest neighbor selection step and the class prediction by the teacher model, which can distort the effectiveness of the TEMI and CE loss, we adopted a strategy involving multiple nearest neighbors on large datasets. We sample $m$ neighbors per image $x$, compute the mean of the teacher model's outputs across these neighbors and use this mean in the loss computation. This approach helps smooth out inaccuracies that may arise from individual neighbor predictions, thereby providing a more reliable and stable basis for training.
    \fi
\end{itemize}

\subsection{Cluster Ensembles}
At the end of stage 1, we obtain a multi-head image clustering model that assigns cluster labels for each example in the training set. Essentially, there are $N$ trained clustering heads, and each head can be regarded as a distinct model. Ensembling is a well-established approach to combine different outputs on a problem. However, directly ensembling the results with a hard voting or soft voting approach is not viable due to the lack of correspondence between the clustering outputs of different heads. In other words, while a clustering head may assign a sample to cluster 1, another head might assign it to cluster 5, with both predictions being correct within their respective clustering space. To obtain a ``consensus clustering", we use the cluster ensembles method by \citet{strehl2002cluster} described below.

Let $\lambda$ represent a cluster label vector, specifying cluster indices for all examples in the dataset. For example, for a dataset of $5$ examples and $3$ clusters, the label vector could look as $\lambda=( 3,3,1,1,2)$. Since there are $N$ clustering heads, we have $N$ different clusterings, represented by $\lambda^{(1)}, \lambda^{(2)}, ... \lambda^{(N)}$. The ``consensus clustering", $\lambda^*$, that we are searching for is a clustering that maximizes the normalized mutual information (NMI) between the consensus clustering and individual clusterings coming from $N$ heads: 

\begin{equation}
    \lambda^* =\underset{\lambda}{\arg \max} \sum_{i=1}^N \phi^\mathrm{NMI}(\lambda, \lambda^{(i)}). 
    \label{eq:cluster_ensemble_objective}
\end{equation}

\noindent The normalized mutual information estimate between two given clusterings is defined as

\begin{equation}
   \phi^\mathrm{NMI}(\lambda^{(a)}, \lambda^{(b)}) = 
   \frac{
   \mathrm{MI}(\lambda^{(a)}, \lambda^{(b)})
   }{
   \sqrt{\mathrm{H}(\lambda^{(a)}) \mathrm{H}(\lambda^{(b)})}},
\end{equation}

\noindent where the numerator is the mutual information between $\lambda^{(a)}$ and  $\lambda^{(b)}$, and the denominator is the square-root of product of ``entropies", which are defined as follows: 

\begin{equation}
  \mathrm{MI}(\lambda^{(a)}, \lambda^{(b)}) =
   \sum_{h=1}^{k^{(a)}} \sum_{l=1}^{k^{(b)}} n_{h,l} \log \left( n \frac{n_{h,l}}{n_h^{(a)} n_l^{(b)}} \right), 
\end{equation}

\begin{equation}
\mathrm{H}(\lambda^{(a)}) = \sum_{h=1}^{k^{(a)}} n_h^{(a)} \log \frac{n_h^{(a)}}{n}.
\end{equation}

% \begin{multline}
%    \phi^\mathrm{NMI}(\lambda^{(a)}, \lambda^{(b)}) = \\ \frac{
%    \sum_{h=1}^{k^{(a)}} \sum_{l=1}^{k^{(b)}} n_{h,l} \log \left( n \frac{n_{h,l}}{n_h^{(a)} n_l^{(b)}} \right)
%    }{\sqrt{( \sum_{h=1}^{k^{(a)}} n_h^{(a)} \log \frac{n_h^{(a)}}{n})
%    (\sum_{l=1}^{k^{(b)}} n_l^{(b)} \log \frac{n_l^{(b)}}{n})}},
% \end{multline}

\noindent Here, $n_{h}^{(a)}$ denotes the number of examples in cluster $h$ of $\lambda^{(a)}$, and $n_{l}^{(b)}$ is the same for cluster $l$ of $\lambda^{(b)}$. $n_{h,l}$ is the number of examples that are both in cluster $h$ of $\lambda^{(a)}$ and in cluster $l$ of $\lambda^{(b)}$. $n$ is the total number of examples, and $k^{(a)}, k^{(b)}$ denote the number of clusters in $\lambda^{(a)}, \lambda^{(b)}$, respectively.
%Therefore, we introduce the concept of using a cluster ensembling method, which is aimed at solving this correspondence problem. 
\citet{strehl2002cluster} propose a hypergraph partitioning-based  method to solve the optimization problem in Equation \eqref{eq:cluster_ensemble_objective}. We use this method to consolidate $N$ clustering outputs into a single consensus clustering output. In our experiments, we show that the consensus clustering significantly outperforms the performances of individual clusterings.

\subsection{Self-Training and Inference}
In the final stage, we use the consensus clustering outputs as pseudo-labels for the training images and train a classification model in a supervised manner. We empirically found that multiple rounds of self-training doesn't improve the performance, thus limit the self-training to only one iteration. The trained classification model is later used during the inference on unseen images.

%% file: sec/04_experiments.tex
\section{Experiments}

\subsection{Datasets}
We train and evaluate our proposed framework on 10 different image classification datasets, namely CIFAR10, CIFAR20, CIFAR100 \cite{krizhevsky2009learning}, STL10 \cite{coates2011analysis}, Food101 \cite{bossard2014food} and ImageNet \cite{deng2009imagenet} with Tiny-ImageNet \cite{chrabaszcz2017downsampled}, ImageNet-50, ImageNet-100 and ImageNet-200 \cite{van2020scan} variants.
CIFAR10, CIFAR20 and CIFAR 100 datasets consist of 50k training images and 10k validation images of size 32x32. CIFAR20 dataset has the same training and validation set with the CIFAR100 dataset, while the 100 classes in the CIFAR100 dataset are mapped to 20 superclasses in CIFAR20. STL10 dataset consists of 5k training images and 8k validation images with size 96x96. Food101 dataset consists of 750 training images and 250 test images per class with 101 classes, and ImageNet dataset have 1,281,167 training images and 50k validation images of varying sizes. Tiny-ImageNet \cite{chrabaszcz2017downsampled} consist of 100k training images and 10k validation images that are resized to 64x64. Other variants ---ImageNet-50, ImageNet-100 and ImageNet-200 \cite{van2020scan}--- have 64,274, 128,545 and 256,558 training images and 2,500, 5,000 and 10,000 validation images, respectively.

\begin{table}[b!]
    \centering
    \resizebox{\linewidth}{!}{\begin{tabular}{llll}
        \toprule
        Methods & NMI(\%) & ACC(\%) & ARI(\%) \\
        \midrule
        Baseline (w/ DINO ViT-B/16) & 85.65 & 75.05 & 65.45 \\
        \texttt{+} (our reproduction) & 87.70 & 79.36 & 71.07 \\
        \texttt{+} (w/ DINOv2 ViT-L/14) & 88.75 \textcolor{ao(english)}{$\uparrow$ 1.05} & 79.44 \textcolor{ao(english)}{$\uparrow$ 0.08} & 71.50 \textcolor{ao(english)}{$\uparrow$ 0.43} \\ % exp206
        \texttt{+} Feat. & 89.58 \textcolor{ao(english)}{$\uparrow$ 0.83} & 79.82 \textcolor{ao(english)}{$\uparrow$ 0.38} & 72.66 \textcolor{ao(english)}{$\uparrow$ 1.16} \\ % exp242
        \texttt{+} BN & 90.10 \textcolor{ao(english)}{$\uparrow$ 0.52} & 81.86 \textcolor{ao(english)}{$\uparrow$ 2.04} & 74.44 \textcolor{ao(english)}{$\uparrow$ 1.78} \\ % exp195
        \textbf{\texttt{+} Adaptive NN (0.3)} & 94.83 \textcolor{ao(english)}{$\uparrow$ 4.73} & 91.00 \textcolor{ao(english)}{$\uparrow$ 9.14} & 87.02 \textcolor{ao(english)}{$\uparrow$ 12.58} \\ % exp193
        \texttt{+} SK & 95.03 \textcolor{ao(english)}{$\uparrow$ 0.20} & 92.60 \textcolor{ao(english)}{$\uparrow$ 1.60} & 88.23 \textcolor{ao(english)}{$\uparrow$ 1.21} \\ % exp192
        \textbf{\texttt{+} TEMI w/ CE. term} & \textbf{95.16} \textcolor{ao(english)}{$\uparrow$ 0.13} & \textbf{93.18} \textcolor{ao(english)}{$\uparrow$ 0.58} & \textbf{88.80} \textcolor{ao(english)}{$\uparrow$ 0.57} \\ % exp187
        \midrule
        $\Delta$ & \textcolor{ao(english)}{\textbf{+6.41}} & \textcolor{ao(english)}{\textbf{13.74}} & \textcolor{ao(english)}{\textbf{+17.30}} \\
        \bottomrule
    \end{tabular}}
    \caption[Ablation study on the components over TEMI baseline.]{\textbf{Ablation study on the proposed improvements over TEMI baseline.} We optimize the unsupervised multi-head image clustering performance with several enhancements in feature space and training process. We highlight the proposed novelties in bold. An adaptive distance threshold of 0.3 is used to select the nearest neighbors. Cumulative performance gains provided by the incremental enhancements on ImageNet-100 dataset are presented in the bottom row.}
    \label{tab:stage_1}
\end{table}

\begin{table*}[t!]
    \centering
    \resizebox{\linewidth}{!}{
    \begin{tabular}{lccccccccc}
        \toprule
        & \multicolumn{4}{c}{TEMI Baseline} & \multicolumn{4}{c}{Our Results} & $\Delta$ \\
        \cmidrule(lr{1em}){2-5} \cmidrule(lr{1em}){6-9} \cmidrule(lr{1em}){10-10}
        Datasets & Backbone & NMI(\%) & ACC(\%) & ARI(\%) & Backbone & NMI(\%) & ACC(\%) & ARI(\%) & ACC(\%)\\
        \midrule
        STL10 & 
        \makecell{DINO \\ (ViT-B/16)} & $96.5\pm0.13$ & $98.5\pm0.04$ & $96.8\pm0.04$ & 
        \makecell{DINOv2 \\ (ViT-B/14)} & $98.90\pm0.06$ & $99.57\pm0.04$ & $99.05\pm0.08$ & \textcolor{ao(english)}{+1.1} \\
        CIFAR10 & 
        \makecell{DINO \\ (ViT-B/16)} & $88.6\pm0.05$ & $94.5\pm0.03$ & $88.5\pm0.08$ & 
        \makecell{DINOv2 \\ (ViT-L/14)} & $97.93\pm0.07$ & $99.27\pm0.04$ & $98.38\pm0.08$ & \textcolor{ao(english)}{+4.8} \\
        CIFAR20 & 
        \makecell{DINO \\ (ViT-B/16)} & $65.4\pm0.45$ & $63.2\pm0.38$ & $48.9\pm0.21$ & 
        \makecell{DINOv2 \\ (ViT-L/14)} & $74.09\pm0.83$ & $67.71\pm1.03$ & $55.83\pm1.09$ & \textcolor{ao(english)}{+4.5} \\
        CIFAR100 & 
        \makecell{DINO \\ (ViT-B/16)} & $76.9\pm0.45$ & $67.1\pm1.30$ & $53.3\pm1.02$ & 
        \makecell{DINOv2 \\ (ViT-L/14)} & $90.60\pm0.05$ & $87.49\pm0.24$ & $80.08\pm0.21$ & \textcolor{ao(english)}{+20.4} \\
        ImageNet-50 & 
        \makecell{MSN \\ (ViT-L/16)} & $88.14\pm0.55$ & $84.87\pm1.16$ & $76.46\pm1.17$ & 
        \makecell{DINOv2 \\ (ViT-L/14)} & $96.81\pm0.11$ & $97.06\pm0.10$ & $94.13\pm0.18$ & \textcolor{ao(english)}{+12.1} \\
        ImageNet-100 & 
        \makecell{MSN \\ (ViT-L/16)} & $88.53\pm0.56$ & $82.86\pm0.73$ & $74.08\pm1.20$ & 
        \makecell{DINOv2 \\ (ViT-L/14)} & $95.44\pm0.09$ & $93.42\pm0.26$ & $89.34\pm0.33$ & \textcolor{ao(english)}{+10.5} \\
        ImageNet-200 & 
        \makecell{MSN \\ (ViT-L/16)} & $86.65\pm0.32$ & $77.96\pm0.71$ & $66.70\pm0.71$ & 
        \makecell{DINOv2 \\ (ViT-L/14)} & $93.54\pm0.09$ & $88.64\pm0.26$ & $82.58\pm0.29$ & \textcolor{ao(english)}{+10.6} \\
        ImageNet & 
        \makecell{MSN \\ (ViT-L/16)} & $82.5$ & $61.56\pm0.28$ & $48.4$ & 
        \makecell{DINOv2 \\ (ViT-L/14)} & $87.73\pm0.03$ & $70.23\pm0.16$ & $59.45\pm0.14$ & \textcolor{ao(english)}{+8.6} \\
        \bottomrule
    \end{tabular}
    }
    \caption[Overall improvements with proposed components over the TEMI baseline on various datasets.]{\textbf{Overall improvements with proposed components over the TEMI baseline on eight benchmark datasets.} We report the mean and standard deviation of 5 independent runs with different seeds as our results and adopted the results of TEMI \cite{adaloglou2023exploring} from the corresponding paper. Our enhancements provide significant improvements over the baseline, detailed in the last column as the delta between the mean accuracy of TEMI and our method. Our training hyper-parameters for each dataset can be found in the supplementary material.}
    \label{tab:stage_1_extensive}
\end{table*}

\subsection{Implementation Details}
Similar to TEMI \cite{adaloglou2023exploring} and other previous works on unsupervised image classification, %for a fair comparison, 
we assume to know the number of ground truth classes/clusters in the datasets during training, which is common practice in the unsupervised image classification literature. 
We resize all images to the size of $224 \times 224$ and use an AdamW optimizer \cite{loshchilov2017fixing} with a weight decay of $10^{-4}$. We used 800 epochs for the STL10 dataset given the fewer number of images in the training set, and 400 epochs for other datasets. We used a batch size of 256 on variants of CIFAR datasets and 1024 on other datasets. % as we found works better. 
Following TEMI, we set the number of classifiers heads $H = 50$.

Similar to TEMI \cite{adaloglou2023exploring}, we found that data augmentations don't improve the performance, thus we precompute the feature representations and reuse them for a faster training and evaluation during unsupervised classifier training and self-training stages. We performed our experiments on a single A100 GPU and found that our extensions result in training times similar to the TEMI baseline for 200 epochs, but require double the training time for 400 epochs with marginal performance improvements. On the ImageNet dataset, stage-1 training takes 24 hours to complete, while stage-2 and stage-3 take approximately 3 hours in total.

For the self-training experiments, we follow DINOv2 \cite{oquab2023dinov2} evaluation protocol to search for the optimal hyperparameters, where we experiment with different learning rates, how many output layers we use, and whether or not to concatenate the average-pooled patch token features with the class token in one training. We train the linear layers for 12500 iterations using an SGD optimizer and report the accuracy of the best performing setting. For  cluster ensembling, we use the public implementation\footnote{\url{https://github.com/GGiecold-zz/Cluster_Ensembles}} of Cluster Ensembles \cite{strehl2002cluster} method.

\subsection{Experimental Results}
We evaluate our models on the validation set of the datasets using Hungarian matching algorithm to map the predicted clusters to the ground truth labels, which is a common practice in unsupervised image classification literature.  We report our results using three main evaluation metrics: clustering accuracy (ACC), normalized mutual information (NMI) and adjusted rand index (ARI).

\subsubsection{Improved Unsupervised Classification Training}
Our stage-1 improves over the TEMI \cite{adaloglou2023exploring} multi-head clustering baseline by adding two main components along with a bag of tricks enhancements described in Section \ref{sec:stage_1}. We train multiple models where we add our proposed components incrementally to showcase the contribution of each component over the baseline. First, we switch from DINO \cite{caron2021emerging} features to DINOv2 \cite{oquab2023dinov2} features to better leverage the enhanced feature quality and improved nearest neighbor performance. Next, we improve the feature space provided by the frozen backbone by adding the output features of the last attention block to the CLS token which slightly improves the nearest neighbor quality. We also switch the embedding normalization layer of TEMI with a batch normalization layer for training stability. We use a similarity threshold of $0.3$ to select nearest neighbors, and a teacher temperature of $0.1$ is used combined with the Sinkhorn-Knopp (SK) normalization. We report the performance metric scores in Table \ref{tab:stage_1} on the validation set of ImageNet-100 dataset. Our proposed components incrementally contribute to the clustering performance, increasing the clustering accuracy of the baseline method from $79.36\%$ to $93.18\%$ with an absolute $13.74$ percentage points, while adaptive nearest neighbor selection approach single-handedly improve the clustering accuracy with an absolute $9.14$ points. Using adaptive nearest neighbor selection method increases the average number of selected nearest neighbors per-sample from $50$ to $940.7$, which attributes for most of the correct pairs from the same semantic label considering the high nearest neighbor performance of DINOv2 features. This number decreases to $651.4$ for a similarity threshold of $0.5$. We introduce the cross-entropy loss term as the final component addition to the baseline because this component works best when the model's performance is at the top due to the usage of the model predictions as pseudo-labels in the cross-entropy computation.

We also provide an extensive comparison of the improved clustering performance with the TEMI baseline based on the proposed components on 8 different unsupervised classification benchmarks in Table \ref{tab:stage_1_extensive}. We report the mean and standard deviation of 5 different runs with random seeds. We show that our improvements consistently improve the TEMI baseline on all of the benchmark datasets, closing the gap between unsupervised and supervised image classification.

\begin{table}[t!]
    \centering
    \resizebox{\linewidth}{!}{
    \begin{tabular}{lcccccc}
        \toprule
        & \multicolumn{3}{c}{w/o Smoothing} & \multicolumn{3}{c}{w/ Smoothing} \\
        \cmidrule(lr{1em}){2-4} \cmidrule(lr{1em}){5-7}
        Datasets & NMI(\%) & ACC(\%) & ARI(\%) & NMI(\%) & ACC(\%) & ARI(\%) \\
        \midrule
        CIFAR20 & 74.72 & 68.38 & 56.61 & 75.46 & 70.63 & 58.44 \\
        CIFAR100 & 90.62 & 87.76 & 80.32 & 90.35 & 88.09 & 79.87 \\
        ImageNet-100 & 95.33 & 92.92 & 88.73 & 95.13 & 92.78 & 88.59 \\
        ImageNet & 87.77 & 70.34 & 59.67 & 87.66 & 71.46 & 59.98 \\
        \bottomrule
    \end{tabular}
    }
    \caption[Effects of multiple nearest neighbors smoothing.]{\textbf{Effects of Multiple Nearest Neighbors Smoothing.} We report the performance metric comparison of the best classifier heads of models trained with and without multiple nearest neighbors smoothing. Corresponding models share the same random number generation seed and are trained with all the proposed components.}
    \label{tab:multiple_smoothing}
\end{table}

\begin{table}[t!]
    \centering
    \resizebox{\linewidth}{!}{
    \begin{tabular}{lcccccc}
        \toprule
        & \multicolumn{3}{c}{Adaptive Nearest Neighbors} & \multicolumn{3}{c}{Ground-Truth Nearest Neighbors} \\
        \cmidrule(lr{1em}){2-4} \cmidrule(lr{1em}){5-7}
        Datasets & NMI(\%) & ACC(\%) & ARI(\%) & NMI(\%) & ACC(\%) & ARI(\%)\\
        \midrule
        STL10 & 
        99.00 & 99.64 & 99.19 &
        99.16 & 99.70 & 99.34 \\
        CIFAR10 & 
        98.03 & 99.32 & 98.49 &
        98.41 & 99.46 & 98.81 \\
        CIFAR20 & 
        75.47 & 70.63 & 58.44 &
        94.10 & 97.08 & 93.97 \\
        CIFAR100 & 
        90.35 & 88.09 & 9.87 &
        93.96 & 92.10 & 87.35 \\
        ImageNet-50 & 
        96.89 & 97.16 & 94.31 &
        98.39 & 98.68 & 97.32 \\
        ImageNet-100 & 
        95.54 & 93.62 & 89.67 &
        96.70 & 96.38 & 93.07 \\
        ImageNet-200 & 
        93.37 & 88.79 & 82.39 &
        96.04 & 94.25 & 89.90 \\
        ImageNet &
        87.66 & 71.46 & 59.98 & 
        91.37 & 81.21 & 71.25 \\
        \bottomrule
    \end{tabular}
    }
    \caption[Upper bound analysis of multi-head clustering based on nearest neighbors quality.]{\textbf{Upper bound analysis of multi-head clustering models based on nearest neighbors quality.} We report the unsupervised image clustering performance of trained models under adaptive nearest neighbors and ground-truth nearest neighbors. Clustering accuracy shows significant improvement across datasets despite the lack of human labels during training.}
    \label{tab:upper_bound}
\end{table}

Finally, we present the multiple neighbors smoothing results on 4 different datasets in Table \ref{tab:multiple_smoothing}. While multiple neighbors smoothing effected the performance on ImageNet-100 dataset slightly negatively, it improves the result over 3 benchmark datasets, improving the ImageNet clustering accuracy from 70.34\% to 71.46\% with a 1.12 point improvement.

\begin{table}[t!]
    \centering
    \resizebox{\linewidth}{!}{
    \begin{tabular}{llll}
        \toprule
        Methods & NMI(\%) & ACC(\%) & ARI(\%) \\
        \midrule
        Baseline (w/ DINOv2 ViT-L/14) & 88.75 & 79.44 & 71.50 \\ % exp206
        \hspace{0.5cm} \texttt{+} Ensemble + Self-Training \textdaggerdbl & 93.17 \textcolor{ao(english)}{$\uparrow$ 4.42} & 88.34 \textcolor{ao(english)}{$\uparrow$ 8.90} & 82.30 \textcolor{ao(english)}{$\uparrow$ 10.80} \\
        \midrule
        \texttt{+} Feat. & 89.58 & 79.82 & 72.66 \\ % exp242
        \hspace{0.5cm} \texttt{+} Ensemble + Self-Training \textdaggerdbl & 93.90 \textcolor{ao(english)}{$\uparrow$ 4.32} & 90.54 \textcolor{ao(english)}{$\uparrow$ 10.72} & 84.74 \textcolor{ao(english)}{$\uparrow$ 12.08} \\
        \midrule
        \texttt{+} BN & 90.10 & 81.86 & 74.44 \\ % exp195
        \hspace{0.5cm} \texttt{+} Ensemble + Self-Training \textdaggerdbl & 94.41 \textcolor{ao(english)}{$\uparrow$ 4.31} & 90.84 \textcolor{ao(english)}{$\uparrow$ 8.98} & 85.34 \textcolor{ao(english)}{$\uparrow$ 10.90} \\
        \midrule
        \texttt{+} Adaptive NN (0.3) & 94.83 & 91.00  & 87.02 \\ % exp193
        \hspace{0.5cm} \texttt{+} Ensemble + Self-Training \textdaggerdbl & 94.99 \textcolor{ao(english)}{$\uparrow$ 0.16} & 92.40 \textcolor{ao(english)}{$\uparrow$ 1.40} & 87.92 \textcolor{ao(english)}{$\uparrow$ 0.90} \\
        \midrule
        \texttt{+} SK & 95.03 & 92.60 & 88.23 \\ % exp192
        \hspace{0.5cm} \texttt{+} Ensemble + Self-Training \textdaggerdbl & 95.06 \textcolor{ao(english)}{$\uparrow$ 0.03} & 92.54 \textcolor{Maroon}{$\downarrow$ 0.06} & 88.19 \textcolor{Maroon}{$\downarrow$ 0.04} \\
        \midrule
        \texttt{+} TEMI w/ CE. & 95.16 & 93.18 & 88.80 \\ % exp187
        \hspace{0.5cm} \texttt{+} Ensemble + Self-Training \textdaggerdbl & 95.20 \textcolor{ao(english)}{$\uparrow$ 0.04} & 93.16 \textcolor{Maroon}{$\downarrow$ 0.02} & 88.80 = \\
        \bottomrule
    \end{tabular}}
    \caption[Cluster ensembling results on different accuracy levels.]{\textbf{Cluster Ensembling Results on Different Accuracy Levels on ImageNet-100 dataset.} Cluster ensembling provides improvements across different accuracy spectrum. While the improvement is substantial for models with lower accuracy, the gap between the baseline and the ensembling narrows for higher accuracy models due to reduced diversity in clustering errors.}
    \label{tab:ensembling}
\end{table}

\noindent \textbf{Upper Bound Analysis with Ground Truth Nearest Neighbors.}
Following significant improvements through various enhancements to the TEMI \cite{adaloglou2023exploring} baseline, we investigate the upper bound for multi-head classifier training based on the nearest neighbor sets. This analysis is intended to underscore the importance of nearest neighbor set quality during unsupervised training. To achieve this, we train new models incorporating all the enhancements proposed in stage-1 of our method using the ground truth nearest neighbors for each sample, defined as samples sharing the same semantic label. Table \ref{tab:upper_bound} presents clustering accuracies on eight benchmark datasets with adaptive nearest neighbor selection and ground truth nearest neighbors.
The results show that clustering accuracy improves significantly across datasets, even without direct human label usage during training, indicating that there remains further potential to enhance the multi-head clustering approach through better nearest neighbor selection methods. In particular, using ground-truth neighbors provides a 26.55-point absolute clustering accuracy improvement on the CIFAR20 dataset, a 20-class variant of CIFAR100 in which 100 classes are merged into 20 super-classes. This effect reflects the nature of CIFAR20, where instances of the same class do not always share visually similar features (such as cloud, forest, mountain and sea being merged into "outdoor scenes" super-class), leading the nearest neighbor selection process to overlook certain relevant neighbors.

\begin{table*}[t!]
    \centering
    \resizebox{\linewidth}{!}{
    \begin{tabular}{lcccccccccc}
        \toprule
        Datasets & CIFAR10 & CIFAR20 & STL10 & CIFAR100 & Food101 & ImageNet-50 & ImageNet-100 & ImageNet-200 & Tiny-ImageNet & ImageNet \\
        Methods & \\
        \midrule
        DCCM \cite{wu2019deep} & 62.3 & 32.7 & 48.2 & -- &  -- & -- & -- & -- & 10.8 & -- \\
        DeepCluster \cite{caron2018deep} & 37.4 & 18.9 & 33.4 & -- & -- & -- & -- & -- & -- & -- \\
        PICA \cite{huang2020deep} & 69.6 & 33.7 & 71.3 & -- & -- & -- & -- & -- & -- & -- \\
        GCC \cite{zhong2021graph} & 85.6 & 47.2 & 78.8 & -- & -- & -- & -- & -- & -- & -- \\
        NNM \cite{dang2021nearest} & 84.3 & 47.7 & 80.8 & -- & -- & -- & -- & -- & -- & -- \\
        PCL \cite{li2020prototypical} & 87.4 & 52.6 & 41.0 & -- & -- & -- & -- & -- & -- & -- \\
        SCAN \cite{van2020scan} & 88.3 & 50.7 & 80.9 & -- & -- & 76.8 & 68.9 & 58.1 & -- & 39.9 \\
        SCAN + RUC \cite{park2021improving} & 90.1 & 54.5 & 86.6 & -- & -- & -- & -- & -- & -- & -- \\
        ProPos* \cite{huang2022learning} & 94.3 & 61.4 & 86.7 & -- & -- & -- & -- & -- & 25.6 & -- \\
        SPICE \cite{niu2022spice} & 88.7 & -- & 86.8 & -- & -- & -- & -- & -- & \underline{30.5} & -- \\
        TCL \cite{li2022twin} & -- & -- & -- & 53.1 & -- & -- & -- & -- & -- & -- \\
        TSP \cite{zhou2022deep} & 94.0 & 55.6 & 97.9 & $55.6\pm2.50$ & -- & -- & -- & -- & -- & -- \\
        CoKe$\S$ \cite{qian2022unsupervised} & 85.7 & 49.7 & -- & -- & -- & -- & -- & -- & -- & 47.6 \\
        SeCu \cite{qian2023stable} & 93.0 & 55.2 & 83.6 & 51.3 & -- & -- & -- & -- & -- & 53.5 \\
        TEMI (w/ DINO ViT-B/16) \cite{adaloglou2023exploring} & $94.5\pm0.03$ & \underline{$63.2\pm0.38$} & \underline{$98.5\pm0.04$} & $67.1\pm1.30$ & -- & $80.01\pm1.26$ & $75.05\pm1.11$ & $73.12\pm0.72$ & -- & 58.4 \\
        TEMI (w/ MSN ViT-L/16) \cite{adaloglou2023exploring} & $90.0\pm0.14$ & $57.8\pm0.42$ & $96.7\pm0.89$ & $61.4\pm0.16$ & -- & \underline{$84.87\pm1.16$} & \underline{$82.86\pm0.73$} & \underline{$77.96\pm0.71$} & -- & 61.6 \\
        MIM-Refiner \cite{alkin2024mim} & -- & -- & -- & -- & -- & -- & -- & -- & -- & 67.4 \\
        TURTLE DINOv2 ViT-g/14 \cite{gadetsky2024let} & \underline{99.3} & -- & 72.3 & \underline{87.1} & \underline{78.9} & -- & -- & -- & -- & \underline{69.1} \\
        %PRO-DSC \cite{mengexploring} & 97.2\pm0.2 & \underline{71.6\pm1.2} & \underline{98.5\pm0.04} & 77.3\pm1.00 & -- & 80.01\pm1.26 & 75.05\pm1.11 & 73.12\pm0.72 & -69.8\pm1.1 & 65.0\pm1.2\\
        \midrule
        \textit{Ours:} & & & & & & & & & & \\
        ICCE stage-1  & $99.27\pm0.04$ & $67.71\pm1.03$ & $99.57\pm0.04$ & $87.49\pm0.24$ & $80.54\pm0.33$ & $97.06\pm0.10$ & $93.42\pm0.26$ & $88.64\pm0.26$ & $84.34\pm0.21$ & $70.23\pm0.16$ \\
        ICCE stage-1 / Best  & \textbf{99.32} & 69.26 & \textbf{99.64} & 87.76 & 80.89 & \textbf{97.16} & 93.62 & 88.79 & \textbf{84.62} & 70.34 \\
        ICCE Ensemble + Self-Training & 99.20 & \textbf{74.23} & 99.58 & \textbf{88.14} & \textbf{81.60} & 96.72 & \textbf{93.66} & \textbf{89.00} & 84.35 & \textbf{70.36} \\
        \bottomrule
    \end{tabular}}
    \caption[Comparison of our method with the state-of-the-art on 10 image clustering datasets.]{\textbf{Comparison with the State-of-the-art on 10 different image classification benchmark datasets.} We report the clustering accuracy with mean and standard deviation of 5 independent runs with different seeds as our stage-1 results. We also report the results of best of five trainings from stage-1. ICCE sets new state-of-the-art results in unsupervised image classification/clustering problem on all 10 benchmark datasets. We use DINOv2 ViT-Large model \cite{darcet2023vision} for all experiments except from STL10 dataset, which we use DINOv2 ViT-Base model. The best results are boldfaced, and the second-best results are underlined. $\S$ results are taken from SeCu \cite{qian2023stable} paper.}
    \label{tab:ucls_sota_comparison3}
\end{table*}

\subsubsection{Cluster Ensembling and Self-Training}
We conduct the ensembling and self-training experiments accross different clustering accuracy levels of the multi-head clustering models to provide insights on how well the ensembling approach can utilize the diverse clustering outputs of varying qualities to improve the performance of the lowest cost clustering head. Table \ref{tab:ensembling} show that, while performance improvements plateau on ImageNet-100 dataset as the performance of the unsupervised classification model improves, ensembling can yield substantial performance gains for lower-performing models. We attribute the diminishing returns in better clustering models to reduced diversity in clustering outputs accross different clustering heads. Ensembling methods primarily rely on diversity in error patterns to enhance performance \cite{dietterich2000ensemble}, and while clustering heads with lower accuracy often make varied errors in cluster assignments, this diversity is reduced in higher-performing clustering heads. Nevertheless, cluster ensembling consistently provides equal or better final performance for all datasets.

\subsubsection{Comparison with the State-of-the-Art}
We present the performance comparison of our method, \textit{ICCE}, with the current state-of-the-art on 10 different unsupervised image classification benchmark datasets in fully unsupervised settings in Table \ref{tab:ucls_sota_comparison3}. We set a new state of the art on unsupervised image classification/clustering problem on all of the benchmark datasets, exceeding 70\% clustering accuracy on ImageNet dataset for the first time in fully unsupervised settings. It should be noted that we exclude CLIP \cite{radford2021learning} backbone results from previous methods in the comparison for the fully unsupervised settings as CLIP pretraining involves image-text pairs as supervision.

%% file: sec/05_conclusion.tex
\section{Conclusion and Discussion}

In this paper, we introduced an enhanced unsupervised image classification framework by proposing several improvements to the multi-head classifier approach. We achieve state-of-the-art results in unsupervised image classification problem on ten image classification benchmarks with large margins. To the best of our knowledge, we are the first to break the 70\% barrier on ImageNet dataset in the fully unsupervised image classification task. We also propose using a cluster ensembling method to better leverage the multiple heads in the classifiers and further enhance our results through a self-training step. We analyze each of the components in our framework in detail through incremental analysis and ablation studies. We provide further ablations in the supplementary material.

There are certain limitations to our proposed unsupervised classification framework. While our stage-1 improvements already achieve state-of-the-art results, the cluster ensembling stage, which provides additional performance improvements, can make the framework cumbersome and not directly applicable to other domains such as unsupervised object detection and instance segmentation, which include both classification and localization branches that require end-to-end training. We evaluate potential improvements to simplify the framework and create an end-to-end unsupervised classification pipeline, aiming to reduce the extra effort required during ensembling and self-training. One consideration is using a codebook during training to align cluster assignments between multiple heads of the classifier. This could enable the use of classifier ensembling methods, such as hard voting or soft voting, and eliminate the need for a separate cluster ensembling stage. Additionally, there is a gap between the upper bound (based on ground truth nearest neighbor experiments presented in the supplementary material) and the current results of ICCE. Better utilization of the nearest neighbor set with possible false positive filtering heuristics can improve the final performance.

\subsection*{Acknowledgements}
We gratefully acknowledge the computational resources
kindly provided by T\"UB\. ITAK ULAKB\. IM High Performance and Grid Computing Center (TRUBA). Dr. Akbas
gratefully acknowledges the support of TUBITAK 2219.

%% file: X_suppl.tex
\clearpage
\setcounter{page}{1}
\maketitlesupplementary

\section{Training Details for Multi-Head Classifiers}
In Table \ref{tab:hyperparams}, we provide the hyper-parameters for training the clustering heads on several benchmark datasets. While our results in Table \ref{tab:stage_1} are obtained by using the default hyper-parameters reported in TEMI \cite{adaloglou2023exploring} paper, we perform further hyper-parameter search for the DINOv2 features that are used in the results of Table \ref{tab:stage_1_extensive} to utilize these features to the full extent. We train our models for 400 epochs which provides equal performance or slight improvements over training with 200 epochs.

\begin{table}[h]
    \centering
    \resizebox{\linewidth}{!}{
    \begin{tabular}{c c}
        \toprule
        config & value \\
        \midrule
        optimizer & AdamW \\
        base learning rate & $1.25e-6, 2.5e-5$ (CIFAR20) \\
        weight decay & $1e-4$ \\
        optimizer momentum & $\beta_1,\beta_2=0.9, 0.999$ \\
        batch size & $1024$ \\
        learning rate schedule & constant \\
        softmax temperature $\tau$ & $0.1, 0.07$ (ImageNet) \\
        $\beta$ & $0.6$, $0.55$ (CIFAR20) \\
        cluster heads & $50$ \\
        warmup epochs & $100$ \\
        total epochs & $400$, $800$ (STL10, CIFAR20) \\
        teacher momentum & $0.996$ \\
        augmentation & None \\
        nearest neighbor threshold & $0.3$ \\
        \bottomrule
    \end{tabular}
    }
    \caption{\textbf{Hyperparameters for training the clustering heads.}}
    \label{tab:hyperparams}
\end{table}

\section{Oracle knowledge of K.} In Table \ref{tab:oracle_k}, we evaluated our method under ±20\% variation in K.  Given the 80\% accuracy upper bound for K=80, these results show that our method remains effective under both overestimation and underestimation.
Oracle knowledge of K is common practice in prior work \cite{adaloglou2023exploring, gadetsky2024let, van2020scan}.
\begin{table}[h!]
    \centering
    \begin{tabular}{|c|c|c|}
    \hline
     K & IN100 &  CIFAR100 \\
     \hline
    80 & 77.62\% & 74.65\%\\ 
    120 & 92.26\% & 88.14\%\\
    \hline
    \end{tabular}
    \caption{\textbf{Clustering Performance Under Different K Values.} We report the accuracy comparison of the best classifier heads for over and under clustering, showing the robustness of our method.}
    \label{tab:oracle_k}
\end{table}

\section{Accuray Comparison with State-of-the-Art}
In Table \ref{tab:sota_comparison}, we present a comparison of our accuracies with the second and third best results across ten image classification benchmarks to facilitate evaluation. Our results demonstrate a substantial improvement over the previous state-of-the-art wherever there was potential for enhancement.

\begin{table}[h]
\centering
\begin{tabular}{l l l l }
\toprule
\textbf{Dataset} & \textbf{\textcolor{gray}{$3^\mathrm{rd}$}} & \textbf{\textcolor{darkgray}{$2^\mathrm{nd}$}} & \textbf{Ours}  \\
\midrule
CIFAR10 & \textcolor{gray}{$94.5$ \cite{adaloglou2023exploring}} & \textcolor{darkgray}{$99.3$ \cite{gadetsky2024let}} & $99.3$ \footnotesize \textcolor{ao(english)}{$\uparrow 0.0$} \\
CIFAR20 & \textcolor{gray}{$61.4$ \cite{huang2022learning}} & \textcolor{darkgray}{$63.2$ \cite{adaloglou2023exploring}} & $74.2$ \footnotesize \textcolor{ao(english)}{$\uparrow 11.0$} \\
CIFAR100 & \textcolor{gray}{$67.1$ \cite{adaloglou2023exploring}} & \textcolor{darkgray}{$87.1$ \cite{gadetsky2024let}} & $88.1$ \footnotesize \textcolor{ao(english)}{$\uparrow 1.0$} \\
STL10 & \textcolor{gray}{$97.9$ \cite{zhou2022deep}} & \textcolor{darkgray}{$98.5$ \cite{adaloglou2023exploring}} & $99.6$ \footnotesize \textcolor{ao(english)}{$\uparrow 1.1$} \\
Food101 & \textcolor{gray}{--} & \textcolor{darkgray}{$78.9$ \cite{gadetsky2024let}} & $81.6$ \footnotesize \textcolor{ao(english)}{$\uparrow 2.7$} \\
ImageNet-50 & \textcolor{gray}{$76.8$ \cite{van2020scan}} & \textcolor{darkgray}{$84.9$ \cite{adaloglou2023exploring}} & $97.2$ \footnotesize \textcolor{ao(english)}{$\uparrow 12.3$} \\
ImageNet-100 & \textcolor{gray}{$68.9$ \cite{van2020scan}} & \textcolor{darkgray}{$82.9$ \cite{adaloglou2023exploring}} & $93.7$ \footnotesize \textcolor{ao(english)}{$\uparrow 10.8$} \\
ImageNet-200 & \textcolor{gray}{$58.1$ \cite{van2020scan}} & \textcolor{darkgray}{$78.0$ \cite{adaloglou2023exploring}} & $89.0$ \footnotesize \textcolor{ao(english)}{$\uparrow 11.0$} \\
Tiny-ImageNet & \textcolor{gray}{$25.6$ \cite{huang2022learning}} & \textcolor{darkgray}{$30.5$ \cite{niu2022spice}} & $84.6$ \footnotesize \textcolor{ao(english)}{$\uparrow 54.1$} \\
ImageNet & \textcolor{gray}{$67.4$ \cite{alkin2024mim}} & \textcolor{darkgray}{$69.1$ \cite{gadetsky2024let}} & $70.4$ \footnotesize \textcolor{ao(english)}{$\uparrow 1.3$} \\

\bottomrule
\end{tabular}
\caption{\textbf{Comparison with SOTA}. The second and third best accuracies from the literature along with our results. Our method ICCE achieves the best unsupervised image classification performance across 10 datasets. The performance improvement (in terms of percentage points) of ICCE over the $2^\mathrm{nd}$ best method is shown in green color. }
\label{tab:sota_comparison}
\end{table}

\section{Validation Set Ensembling}
We dig deeper into the cluster ensembling approach and analyze the effects of ensembling in the case where the entire validation set is provided beforehand. We first cluster the validation set images into clusters using the multi-head classifiers (Stage-1), and later do ensembling on the clustering outputs of these classifiers to obtain a consensus clustering of the validation set. It should be noted that this is a special case that is not possible in the case where validation images are provided as a stream and not as a whole. We show in Table \ref{tab:val_ensembles} that validation set ensembling results are not very different from training set ensembling + self-training approach, and can improve the results of Stage-1 multi-head classifier outputs.

\begin{table*}[t!]
    \centering
    \resizebox{\linewidth}{!}{
    \begin{tabular}{lcccccccccc}
        \toprule
        Datasets & CIFAR10 & CIFAR20 & STL10 & CIFAR100 & Food101 & ImageNet-50 & ImageNet-100 & ImageNet-200 & Tiny-ImageNet & ImageNet \\
        Methods & \\
        \midrule
        ICCE stage-1  & $99.27\pm0.04$ & $67.71\pm1.03$ & $99.57\pm0.04$ & $87.49\pm0.24$ & $80.54\pm0.33$ & $97.06\pm0.10$ & $93.42\pm0.26$ & $88.64\pm0.26$ & $84.34\pm0.21$ & $70.23\pm0.16$ \\
        ICCE Ensemble + Self-Training & 99.20 & 74.23 & 99.58 & 88.14 & 81.60 & 96.72 & 93.66 & 89.00 & 84.35 & 70.36 \\
        \midrule
        \textbf{ICCE Ensemble w/ Validation Set} \textdaggerdbl & 99.31 & 73.67 & 99.60 & 88.40 & 81.01 & 97.12 & 93.68 & 88.67 & 84.93 & 70.92 \\
        \bottomrule
    \end{tabular}}
    \caption{\textbf{Comparison of validation set cluster ensembling with main results on 10 different image classification benchmark datasets.} We report the validation set cluster ensembling accuracy along with the Stage-1 multi-head classifier results and ICCE Ensembling and self-training results. Validation set cluster ensembling provide very similar results to training set cluster ensembling + self-training results, providing improvements over the base multi-head classifier outputs. We use DINOv2 ViT-L/14 model \cite{darcet2023vision} for all experiments except from STL10 dataset, which we use DINOv2 ViT-Base model. \textdaggerdbl corresponds to the case where the entire validation set is provided before the inference starts.}
    \label{tab:val_ensembles}
\end{table*}

\section{Nearest Neighbor Threshold Analysis}
\begin{figure*}[h]
\centering
\includegraphics[width=\linewidth]{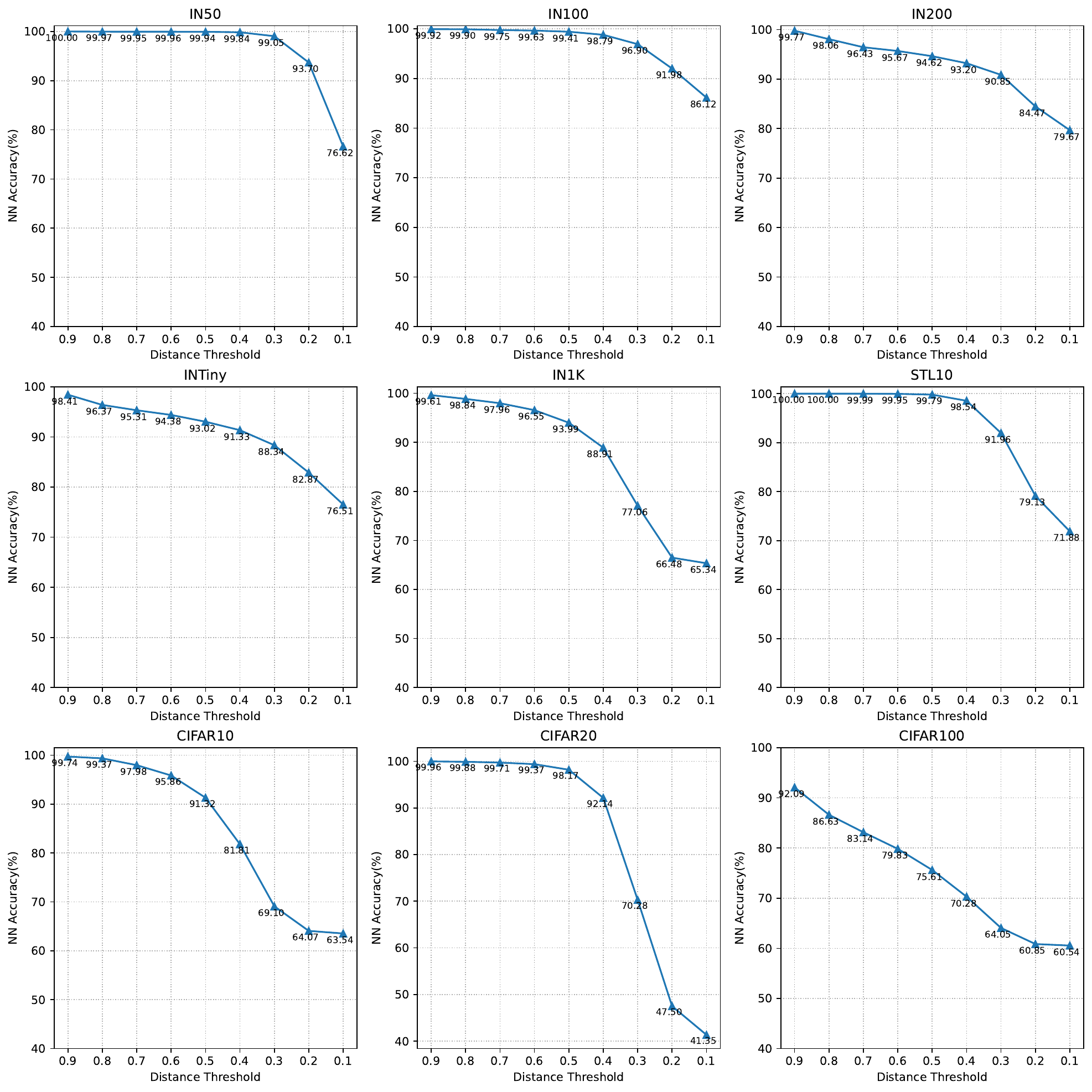}
\caption[Nearest neighbor accuracy analysis on various datasets with DINOv2 ViT-L/14.]{\textbf{Nearest Neighbor Accuracy Analysis on Various Datasets with DINOv2 ViT-L/14.} 
Nearest neighbor accuracy remains very high across all datasets when the distance threshold is set to a higher value. Although the NN accuracy decreases with lower distance threshold values, it remains sufficiently high to achieve performance improvements when used in the nearest neighbor selection process. Best viewed when zoomed in.}
\label{fig:nn_accuracy}
\end{figure*}

We perform nearest neighbor distance threshold analysis on 9 benchmark datasets using DINOv2- ViT-L/14 model to observe the nearest neighbor quality of the backbone model used for training multi-head classifiers. Mainly, we apply different distance thresholds to filter the nearest neighbors set for each image, and compute the accuracy of the nearest neighbor sets based on correct semantic class matches among nearest neighbor pairs. Figure \ref{fig:nn_accuracy} illustrates that the quality of the nearest neighbor sets can raise around the ground truth level for some datasets with correct distance thresholds, and lower threshold still provide a good enough nearest neighbor pair accuracy to be used during the training of the multi-head classifiers.

We further investigate the effects of adaptive distance threshold by training multi-head classifiers on the ImageNet-100 dataset using 10 different equally spaced thresholds, ranging from 0.1 to 1.0, where a threshold of 1.0 corresponds to using only the top-50 nearest neighbors. Table \ref{tab:adaptive_thresholds} demonstrates that a very low threshold of 0.1 significantly outperforms using only the top-50 neighbors, highlighting the effectiveness of the DINOv2 features. Overall mean and standard deviation values show that a distance threshold of 0.3 maximizes the performance across all classifier heads compared to other thresholds.

\begin{table*}[t!]
    \centering
    \resizebox{\linewidth}{!}{
    \begin{tabular}{lcccccccc}
        \toprule
        & \multicolumn{3}{c}{Best Head} & \multicolumn{3}{c}{Overall} & \multicolumn{2}{c}{Nearest Neighbor Stats} \\
        \cmidrule(lr{1em}){2-4} \cmidrule(lr{1em}){5-7} \cmidrule(lr{1em}){8-9}
        Dist. Thr. & NMI(\%) & ACC(\%) & ARI(\%) & NMI(\%) & ACC(\%) & ARI(\%) & Average NN Count & NN ACC(\%) \\
        \midrule
        1.0 & 93.68 & 87.26 & 83.22 & $93.47\pm0.41$ & $87.15\pm1.39$ & $82.51\pm1.33$ & 50.0 & - \\
        0.9 & 94.26 & 89.06 & 84.98 & $93.47\pm0.49$ & $87.37\pm1.52$ & $82.67\pm1.61$ & 71.9 & 99.77 \\
        0.8 & 93.97 & 89.00 & 84.42 & $93.78\pm0.39$ & $88.21\pm1.31$ & $83.60\pm1.32$ & 216.5 & 98.06 \\
        0.7 & 94.69 & 90.56 & 86.28 & $94.01\pm0.43$ & $88.52\pm1.34$ & $84.16\pm1.39$ & 374.9 & 96.43 \\
        0.6 & 94.65 & 91.20 & 86.49 & $94.29\pm0.36$ & $89.22\pm1.28$ & $85.05\pm1.23$ & 511.3 & 95.67 \\
        0.5 & 95.48 & 92.62 & 89.04 & $94.63\pm0.32$ & $90.16\pm1.10$ & $86.12\pm1.13$ & 651.4 & 94.62 \\
        0.4 & 95.37 & 92.20 & 88.57 & $95.09\pm0.22$ & $91.18\pm0.80$ & $87.40\pm0.77$ & 782.6 & 93.20 \\
        0.3 & 95.39 & 92.34 & 88.52 & $95.22\pm0.20$ & $91.83\pm0.81$ & $87.92\pm0.76$ & 940.7 & 90.85 \\
        0.2 & 95.36 & 92.30 & 88.49 & $95.12\pm0.15$ & $91.34\pm0.75$ & $87.41\pm0.64$ & 1158.7 & 84.47 \\
        0.1 & 95.06 & 92.58 & 88.11 & $94.74\pm0.14$ & $90.69\pm0.75$ & $86.48\pm0.66$ & 1284.6 & 79.67 \\
        \bottomrule
    \end{tabular}
    }
    \caption[Ablation study on adaptive nearest neighbor selection distance threshold.]{\textbf{Ablation Study on Adaptive Nearest Neighbor Selection Distance Threshold.} We analyze the effects of adaptive nearest neighbor selection through distance threshold on the performance of multi-head classifiers. The \textit{Best Head} results correspond to the classifier head with the lowest training loss, while the \textit{Overall} results present the mean and standard deviation across all 50 classification heads. We also provide the average number of neighbors per image that are utilized during training, and the nearest neighbor accuracy at the distance threshold over the ImageNet-100 dataset. We use DINOv2-L/14 features for all the experiments. While the classifier with the 0.1 distance threshold provide the single best-head performance, the threshold of 0.3 better propagates the improvements to all the classifier heads.}
    \label{tab:adaptive_thresholds}
\end{table*}

\section{Effect of Number of Classifier Heads}
In TEMI loss function, multi-head classifiers are utilized during the computation of training loss by assigning a low weight to training samples that are likely noisy nearest neighbors. We also use all the classifier heads during the cluster ensembling step to improve the performance of the unsupervised classifiers. To evaluate the effects of number of classifier heads on the multi-head classifier performance, we train several classifier models with number of heads ranging from 10 to 80. As shown in Table \ref{tab:num_heads}, interestingly, classifier with 10 heads performs best on single-head performance, while it drops behind for the ensembling results. Ensembling benefits from the increased number of heads as there are more possibilities for different types of errors which is known to improve ensembling performance \cite{dietterich2000ensemble}. However, we also observed 10-head classifiers train $\sim$5x faster than the 50-head classifiers, providing a strong alternative in limited resources.

\begin{table*}[t!]
    \centering
    \resizebox{\linewidth}{!}{
        \begin{tabular}{lccccccccc}
            \toprule
            & \multicolumn{3}{c}{Best Head} & \multicolumn{3}{c}{Overall} & \multicolumn{3}{c}{Val. Ensembles} \\
            \cmidrule(lr{1em}){2-4} \cmidrule(lr{1em}){5-7}  \cmidrule(lr{1em}){8-10}
            Classifier Heads & NMI(\%) & ACC(\%) & ARI(\%) & NMI(\%) & ACC(\%) & ARI(\%) & NMI(\%) & ACC(\%) & ARI(\%) \\
            \midrule
            10 & 95.37 & 92.54 & 88.64 & $95.07\pm0.14$ & $91.57\pm0.57$ & $87.56\pm0.54$ & 95.23 & 92.42 & 88.38 \\
            20 & 95.27 & 92.74 & 88.54 & $95.07\pm0.18$ & $91.59\pm0.67$ & $87.54\pm0.65$ & 95.26 & 92.54 & 88.35 \\
            30 & 95.22 & 91.88 & 88.09 & $95.16\pm0.19$ & $91.69\pm0.75$ & $87.74\pm0.74$ & 95.50 & 93.40 & 89.38 \\
            40 & 95.41 & 92.14 & 88.50 & $95.20\pm0.21$ & $91.67\pm0.80$ & $87.80\pm0.75$ & 95.41 & 92.74 & 88.69 \\
            50 & 95.39 & 92.34 & 88.52 & $95.22\pm0.20$ & $91.83\pm0.81$ & $87.92\pm0.76$ & 95.40 & 93.32 & 89.19 \\
            60 & 95.48 & 92.20 & 88.62 & $95.17\pm0.22$ & $91.63\pm0.82$ & $87.75\pm0.80$ & 95.63 & 93.70 & 89.82 \\
            70 & 95.42 & 92.08 & 88.49 & $95.21\pm0.20$ & $91.76\pm0.80$ & $87.87\pm0.78$ & 95.50 & 93.42 & 89.38 \\
            80 & 95.28 & 92.04 & 88.12 & $95.21\pm0.18$ & $91.77\pm0.79$ & $87.88\pm0.71$ & 95.61 & 93.60 & 89.72 \\
            \bottomrule
        \end{tabular}
    }
    \caption[Ablation study on the number of classifier heads.]{\textbf{Ablation Study on the Number of Classifier Heads.} We analyze the effect of number of classifier heads to the unsupervised image classification performance. The \textit{Best Head} results correspond to the classifier head with the lowest training loss, while the \textit{Overall} results present the mean and standard deviation across all classification heads. We also provide the cluster ensembling results on the validation set as it is a good approximation of the self-training results based on training set pseudo-labels. We use the ImageNet-100 dataset and DINOv2-L/14 features for the experiments.}
    \label{tab:num_heads}
\end{table*}

\section{Detailed Results on Benchmark Datasets}
We also provide NMI and ARI results along with the clustering accuracy in Table \ref{tab:sota_small}, Table \ref{tab:sota_IN_subset}, Table \ref{tab:sota_big} and Table \ref{tab:sota_food101} on small datasets, ImageNet subsets, Tiny-ImageNet-CIFAR100-ImageNet datasets, and Food 101 dataset, respectively.

\section{Randomly Sampled Images from Cluster
Assignments}
\begin{figure*}[h]
\centering
\includegraphics[width=\linewidth]{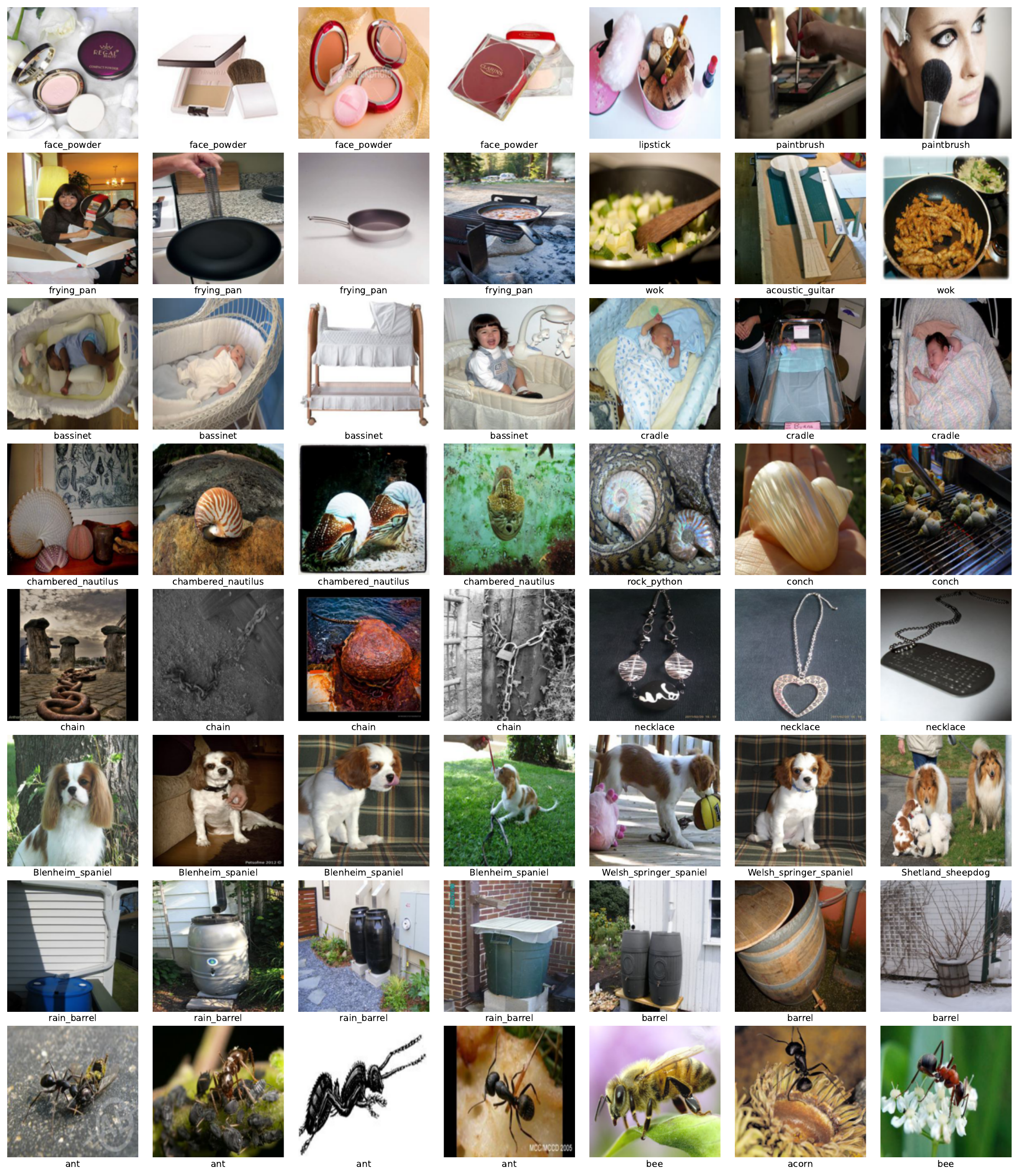}
\caption[Cluster predictions for images from randomly selected classes in the ImageNet dataset.]{\textbf{Cluster predictions for images from randomly selected classes in the ImageNet dataset.} 
Each row contains images that have been assigned to the same cluster by ICCE. The ground-truth label is displayed in the text below each image. The first four columns represent correctly classified images, while the last three columns correspond to misclassified ones.}
\label{fig:true_false}
\end{figure*}

In Figure \ref{fig:true_false}, we illustrate the cluster assignments generated by ICCE on images from the ImageNet dataset. Our observations indicate that, in most cases, the incorrectly assigned clusters are closely related to the correct classes.

\begin{table*}[t!]
    \centering
    \resizebox{\linewidth}{!}{
    \begin{tabular}{lcccccccccc}
        \toprule
        Datasets & \multicolumn{3}{c}{CIFAR10} & \multicolumn{3}{c}{CIFAR20} & \multicolumn{3}{c}{STL10} \\
        \cmidrule(lr{1em}){2-4} \cmidrule(lr{1em}){5-7} \cmidrule(lr{1em}){8-10}
        Methods & NMI(\%) & ACC(\%) & ARI(\%) & NMI(\%) & ACC(\%) & ARI(\%) & NMI(\%) & ACC(\%) & ARI(\%) \\
        \midrule
        DCCM \cite{wu2019deep} & 49.6 & 62.3 & 40.8 & 28.5 & 32.7 & 17.3 & 37.6 & 48.2 & 26.2 \\
        DeepCluster \cite{caron2018deep} & - & 37.4 & - & - & 18.9 & - & - & 33.4 & - \\
        PICA \cite{huang2020deep} & 59.1 & 69.6 & 51.2 & 31 & 33.7 & 17.1 & 61.1 & 71.3 & 53.1 \\
        GCC \cite{zhong2021graph} & 76.4 & 85.6 & 72.8 & 47.2 & 47.2 & 30.5 & 68.4 & 78.8 & 63.1 \\
        NNM \cite{dang2021nearest} & 74.8 & 84.3 & 70.9 & 48.4 & 47.7 & 31.6 & 69.4 & 80.8 & 65 \\
        PCL \cite{li2020prototypical} & 80.2 & 87.4 & 76.6 & 52.8 & 52.6 & 36.3 & 71.8 & 41.0 & 67.0 \\
        SCAN \cite{van2020scan} & 79.7 & 88.3 & 77.2 & 48.6 & 50.7 & 33.3 & 69.8 & 80.9 & 64.6 \\
        SCAN + RUC \cite{park2021improving} & - & 90.1 & - & - & 54.5 & - & - & 86.6 & - \\
        SPICE \cite{niu2022spice} & 86.5 & 92.6 & 85.2 & 56.7 & 53.8 & 38.7 & 87.2 & 93.8 & 87.0 \\
        ProPos* \cite{huang2022learning} & 88.6 & 94.3 & 88.4 & 60.6 & 61.4 & 45.1 & 75.8 & 86.7 & 73.7 \\
        TCL \cite{li2022twin} & 81.9 & 88.7 & 78.0 & - & - & - & 79.9 & 86.8 & 75.7 \\
        TSP \cite{zhou2022deep} & 88.0 & 94.0 & 87.5 & 61.4 & 55.6 & 43.3 & 95.8 & 97.9 & 95.6 \\
        CoKe \cite{qian2022unsupervised} & 76.6 & 85.7 & 73.2 & 49.1 & 49.7 & 33.5 & - & - & - \\
        SeCu \cite{qian2023stable} & 86.1 & 93.0 & 85.7 & 55.1 & 55.2 & 39.7 & 73.3 & 83.6 & 69.3 \\
        TEMI DINO ViT-B/16 \cite{adaloglou2023exploring} & $88.6\pm0.05$ & $94.5\pm0.03$ & $88.5\pm0.08$ & $65.4\pm0.45$ & $63.2\pm0.38$ & $48.9\pm0.21$ & $96.5\pm0.13$ & $98.5\pm0.04$ & $96.8\pm0.09$ \\
        TEMI MSN ViT-L/16 \cite{adaloglou2023exploring} & $82.9\pm0.16$ & $90.0\pm0.14$ & $80.7\pm0.22$ & $59.8\pm0.04$ & $57.8\pm0.42$ & $42.5\pm0.08$ & $93.6\pm1.10$ & $96.7\pm0.89$ & $93.0\pm1.74$ \\
        TURTLE DINOv2 ViT-g/14 \cite{gadetsky2024let} & - & 99.3 & - & - & - & - & - & 72.3 & - \\
        \midrule
        \textit{Ours:} & & & & & & & & & & \\
        ICCE stage-1 & $97.93\pm0.07$ & $99.27\pm0.04$ & $98.38\pm0.08$ & $74.09\pm0.83$ & $67.71\pm1.03$ & $55.83\pm1.09$ & $98.90\pm0.06$ & $99.57\pm0.04$ & $99.05\pm0.08$ \\
        ICCE stage-1 / Best & 98.03 & \textbf{99.32} & \textbf{98.50} & 75.38 & 69.26 & 57.50 & 99.00 & \textbf{99.64} & \textbf{99.20} \\
        ICCE Ensemble \textdagger & 98.03 & 99.31 & 98.48 & 76.61 & 73.67 & 59.67 & 98.98 & 99.60 & 99.12 \\
        ICCE Ensemble + Self-Training \textdaggerdbl & \textbf{98.23} & 99.20 & 97.72 & \textbf{76.49} & \textbf{74.23} & \textbf{60.09} & \textbf{99.06} & 99.58 & 98.88 \\
        \bottomrule
    \end{tabular}
    }
    \caption[Comparison of our method with the state-of-the-art on small-scale datasets.]{\textbf{Comparison with the State-of-the-art on Small-Scale Datasets.} We report the mean and standard deviation of 5 independent runs with different seeds as our stage-1 results. We also report the best results from stage-1. \textdagger corresponds to ensembling on the validation set clustering outputs, while \textdaggerdbl corresponds to ensembling on the training set clustering outputs first and then using self-training on the pseudo-labels obtained from the training set for inference. * indicates methods that utilize validation split during training. We use DINOv2 ViT-Large model \cite{darcet2023vision} for the CIFAR10 and CIFAR20 datasets and DINOv2 ViT-Base \cite{darcet2023vision} model for the STL10 dataset. Best results are boldfaced.}
    \label{tab:sota_small}
\end{table*}

\begin{table*}[h!]
    \centering
    \resizebox{\linewidth}{!}{
    \begin{tabular}{lcccccccccc}
        \toprule
        Datasets & \multicolumn{3}{c}{ImageNet-50} & \multicolumn{3}{c}{ImageNet-100} & \multicolumn{3}{c}{ImageNet-200} \\
        \cmidrule(lr{1em}){2-4} \cmidrule(lr{1em}){5-7} \cmidrule(lr{1em}){8-10}
        Methods & NMI(\%) & ACC(\%) & ARI(\%) & NMI(\%) & ACC(\%) & ARI(\%) & NMI(\%) & ACC(\%) & ARI(\%) \\
        \midrule
        SCAN \cite{van2020scan} & 82.2 & 76.8 & 66.1 & 80.8 & 68.9 & 57.6 & 77.2 & 58.1 & 47.0 \\
        ProPos \cite{huang2022learning} & 82.8 & - & 69.1 & 83.5 & - & 63.5 & 80.6 & - & 53.8 \\
        TEMI MSN ViT-L/16 \cite{adaloglou2023exploring} & $88.14\pm0.55$ & $84.87\pm1.16$ & $76.46\pm1.17$ & $88.53\pm0.56$ & $82.86\pm0.73$ & $74.08\pm1.20$ & 8$6.65\pm0.32$ & $77.96\pm0.71$ & $66.70\pm0.71$ \\
        \midrule
        \textit{Ours:} & & & & & & & & & & \\
        ICCE stage-1 & $96.81\pm0.11$ & $97.06\pm0.10$ & $94.13\pm0.18$ & $95.44\pm0.09$ & $93.42\pm0.26$ & $89.34\pm0.33$ & $93.54\pm0.09$ & $88.64\pm0.26$ & $82.58\pm0.29$ \\
        ICCE stage-1 / Best & \textbf{96.95} & \textbf{97.16} & \textbf{94.33} & 95.54 & 93.62 & 89.67 & 93.37 & 88.79 & 82.40 \\
        ICCE Ensemble \textdagger & 96.87 & 97.12 & 94.24 & 95.56 & 93.68 & 89.71 & 93.36 & 88.67 & 82.29 \\
        ICCE Ensemble + Self-Training \textdaggerdbl & 96.44 & 96.72 & 93.48 & \textbf{95.55} & \textbf{93.66} & \textbf{89.67} & \textbf{93.43} & \textbf{89.00} & \textbf{82.58} \\
        \bottomrule
    \end{tabular}
    }
    \caption[Comparison of our method with the state-of-the-art on ImageNet subsets.]{\textbf{Comparison with the State-of-the-art on ImageNet Subsets.} We report the mean and standard deviation of 5 independent runs with different seeds as our stage-1 results. We also report the best results from stage-1. \textdagger corresponds to ensembling on the validation set clustering outputs, while \textdaggerdbl corresponds to ensembling on the training set clustering outputs first and then using self-training on the pseudo-labels obtained from the training set for inference. We use DINOv2 ViT-Large model \cite{darcet2023vision} for our experiments. Best results are boldfaced.}
    \label{tab:sota_IN_subset}
\end{table*}

\begin{table*}[t!]
    \centering
    \resizebox{\textwidth}{!}{
    \begin{tabular}{lcccccccccc}
        \toprule
        Datasets & \multicolumn{3}{c}{CIFAR100} & \multicolumn{3}{c}{Tiny-ImageNet} & \multicolumn{3}{c}{ImageNet} \\
        \cmidrule(lr{1em}){2-4} \cmidrule(lr{1em}){5-7} \cmidrule(lr{1em}){8-10}
        Methods & NMI(\%) & ACC(\%) & ARI(\%) & NMI(\%) & ACC(\%) & ARI(\%) & NMI(\%) & ACC(\%) & ARI(\%) \\
        \midrule
        DCCM \cite{wu2019deep} & - & - & - & 22.4 & 10.8 & 3.8 & - & - & - \\
        SCAN \cite{van2020scan} & - & - & - & - & - & - & 72.0 & 39.9 & 27.5 \\
        ProPos \cite{huang2022learning} & - & - & - & 40.5 & 25.6 & 14.3 & - & - & - \\
        SPICE \cite{niu2022spice} & - & - & - & 44.9 & 30.5 & 16.1 & - & - & - \\
        TCL \cite{li2022twin} & 52.9 & 53.1 & 35.7 & - & - & - & - & - & - \\
        TSP \cite{zhou2022deep} & $61.4\pm1.4$ &  $55.6\pm2.5$ & $43.3\pm1.8$ & - & - & - & - & - & - \\
        CoKe$\S$ \cite{qian2022unsupervised} & - & - & - & - & - & - & 76.2 & 47.6 & 35.6 \\
        SeCu \cite{qian2023stable} & 65.2 & 51.3 & 37.1 & - & - & - & 79.4 & 53.5 & 41.9 \\
        TEMI DINO ViT-B/16 \cite{adaloglou2023exploring} & $76.9\pm0.45$ & $67.1\pm1.30$ & $53.3\pm1.02$ & - & - & - & 82.5 & $61.56\pm0.28$ & 48.4 \\
        MIM-Refiner \cite{alkin2024mim} & - & - & - & - & - & - & 86.3 & 67.4 & 40.5 \\
        TURTLE DINOv2 ViT-g/14 \cite{gadetsky2024let} & - & 87.1 & - & - & - & - & - & 69.1 & - \\
        \midrule
        \textit{Ours:} & & & & & & & & & & \\
        ICCE stage-1 & $90.60\pm0.05$ & $87.49\pm0.24$ & $80.08\pm0.21$ & $89.93\pm0.10$ & $84.34\pm0.21$ & $74.55\pm0.32$ & $87.73\pm0.03$ & $70.23\pm0.16$ & $59.45\pm0.14$ \\
        ICCE stage-1 / Best & 90.62 & 87.76 & \textbf{80.32} & \textbf{90.03} & \textbf{84.62} & \textbf{75.00} & \textbf{87.77} & 70.34 & \textbf{59.67} \\
        ICCE Ensemble \textdagger & 90.97 & 88.40 & 80.75 & 90.11 & 84.93 & 75.18 & 87.84 & 70.92 & 60.05 \\
        ICCE Ensemble + Self-Training \textdaggerdbl & \textbf{90.65} & \textbf{88.14} & 80.28 & 89.37 & 84.35 & 73.89 & 87.55 & \textbf{70.36} & 58.64 \\
        \bottomrule
    \end{tabular}
    }
    \caption[Comparison of our method with the state-of-the-art on CIFAR100, Tiny-ImageNet and ImageNet.]{\textbf{Comparison with the State-of-the-art on CIFAR100, Tiny-ImageNet and ImageNet.} We report the mean and standard deviation of 5 independent runs with different seeds as our stage-1 results. We also report the best results from stage-1. \textdagger corresponds to ensembling on the validation set clustering outputs, while \textdaggerdbl corresponds to ensembling on the training set clustering outputs first and then using self-training on the pseudo-labels obtained from the training set for inference. We use DINOv2 ViT-Large model \cite{darcet2023vision} for our experiments. Best results are boldfaced. $\S$ results are taken from SeCu \cite{qian2023stable} paper.}
    \label{tab:sota_big}
\end{table*}

\begin{table}[t!]
    \centering
    \resizebox{\linewidth}{!}{
    \begin{tabular}{lccc}
        \toprule
        Dataset & \multicolumn{3}{c}{Food101} \\
        \cmidrule(lr{1em}){2-4}
        Methods & NMI(\%) & ACC(\%) & ARI(\%) \\
        \midrule
        TURTLE DINOv2 ViT-g/14 \cite{gadetsky2024let} & - & 78.9 & - \\
        \midrule
        ICCE stage-1 (Ours) & $85.67\pm0.10$ & $80.54\pm0.33$ & $71.03\pm0.21$ \\
        ICCE stage-1 / Best (Ours) & 85.82 & 80.89 & 71.26 \\
        ICCE Ensemble (Ours)\textdagger & 85.76 & 81.01 & 71.39 \\
        ICCE Ensemble + Self-Training (Ours) \textdaggerdbl & 86.32 & 81.60 & 72.29 \\
        \bottomrule
    \end{tabular}
    }
    \caption[Comparison of our method with the state-of-the-art on Food101 dataset.]{\textbf{Comparison with the State-of-the-art on Food101 Dataset.} We report the mean and standard deviation of 5 independent runs with different seeds as our stage-1 results. We also report the best results from stage-1. \textdagger corresponds to ensembling on the validation set clustering outputs, while \textdaggerdbl corresponds to ensembling on the training set clustering outputs first and then using self-training on the pseudo-labels obtained from the training set for inference. We use DINOv2 ViT-Large model \cite{darcet2023vision} for our experiments.}
    \label{tab:sota_food101}
\end{table}

